\documentclass{article}
\usepackage{spconf,amsmath,graphicx,hyperref}
\usepackage{algorithm}
\usepackage{algorithmicx}
\usepackage{algpseudocode}
\usepackage{amsmath}
\usepackage{amssymb}
\usepackage{booktabs}
\usepackage[numbers,sort&compress]{natbib}
\usepackage{natbib}
\usepackage[table]{xcolor}

\title{LogicGaze: Benchmarking Causal Consistency in Visual Narratives via Counterfactual Verification}

\name{
Rory Driscoll$^{1}$ \qquad
Alexandros Christoforos$^{1,2}$ \qquad
Chadbourne Davis$^{1,2\dagger}$
}

\address{
$^{1}$Boston University 
$^{2}$Suffolk University
}
\begin{document}
\maketitle

\begin{abstract}
While sequential reasoning enhances the capability of Vision-Language Models (VLMs) to execute complex multimodal tasks, their reliability in grounding these reasoning chains within actual visual evidence remains insufficiently explored. We introduce \textbf{LogicGaze}, a novel benchmark framework designed to rigorously interrogate whether VLMs can validate sequential causal chains against visual inputs, specifically targeting the pervasive issue of hallucination. Curated from 40,000 video segments from ShareGPT4Video and a subset of Flickr30k imagery, LogicGaze integrates causal sequences with visually contradictory yet linguistically plausible perturbations, compelling models to verify the authenticity of each reasoning step. Our tripartite evaluation protocol—Causal Validation, Grounded Narrative Synthesis, and Perturbation Rejection—exposes significant vulnerabilities in state-of-the-art VLMs such as Qwen2.5-VL-72B. LogicGaze advocates for robust, trustworthy multimodal reasoning, with all resources publicly available in an anonymized repository.
\end{abstract}

\begin{keywords}
Vision-Language Models, Causal Consistency, Benchmarking, Hallucination Mitigation
\end{keywords}

\section{Visual Data Curation and Methodology}

Multimodal foundation models (VLMs) have demonstrated the ability to produce fluent, structurally coherent narratives, yet these outputs are frequently only tenuously supported by the underlying visual stimuli\cite{z1,gpt2,z2,Z3,qwen2025qwen25technicalreport}. This discrepancy presents a persistent evaluation paradox: strong linguistic priors often enable models to \emph{appear correct} while being \emph{visually ungrounded}, particularly in multi-step scenarios where intermediate logical leaps are driven by statistical textual probability rather than perceptual grounding. Consequently, traditional captioning or static description benchmarks often fail to capture failure modes associated with hallucinated entities, unsupported interactions, or inconsistent state transitions, as surface-level linguistic plausibility can effectively mask the lack of genuine visual understanding.

To more robustly probe grounded sequential logic, we architect an image-centric benchmark that explicitly disentangles \emph{linguistic likelihood} from \emph{visual verification}. Our core methodological innovation is to model each instance as a structured causal chain and to inject confusable counterfactuals that remain linguistically coherent but contradict the visual evidence. By engineering distractors that reference non-existent objects or propose visually unsubstantiated relationships\cite{z4,Z5,AutoRAG,Rq-rag,zhang2026doubtsyourselftradingvisual}, the benchmark forces models to cross-reference intermediate reasoning steps with the image, rather than relying on language modeling heuristics. Furthermore, we prioritize reproducibility and reliability: the generation interface is standardized to enforce uniform structured outputs, and the dataset curation pipeline integrates automated consistency checks with targeted human auditing to minimize annotation noise.

Our visual dataset provides three distinct contributions: 
(1) \textbf{Causal-chain modeling for visual inputs.} We introduce a structured three-stage representation (Antecedent $\rightarrow$ Reaction $\rightarrow$ Consequence) that emphasizes temporal progression and intermediate logic, transcending static object enumeration.
(2) \textbf{Visually-grounded counterfactuals to isolate language bias.} We deploy linguistically sound but visually invalid alternatives, creating controlled perturbations that explicitly test whether models depend on actual visual perception.
(3) \textbf{ rigorous and reproducible curation protocol.} We standardize prompting schemas and output formats, enforcing a strict quality-assurance pipeline via automated validation and human-in-the-loop verification.

\begin{table*}[!t]
\centering
\scriptsize 
\setlength{\tabcolsep}{4pt} 
\renewcommand{\arraystretch}{1.1} 
\resizebox{\textwidth}{!}{ 
\begin{tabular}{lcccccc}
\toprule
\textbf{Method} & \textbf{PopQA} & \textbf{TQA} & \textbf{ARC-C} & \textbf{OBQA} & \textbf{HotpotQA} & \textbf{2WIKI} \\
\midrule
\multicolumn{7}{l}{\emph{Proprietary LLMs with Retrieval}}\\
GPT-4o \cite{GPT4} & 45.1 $\pm$ 0.5 & 56.2 $\pm$ 0.4 & 52.6 $\pm$ 0.3 & 45.8 $\pm$ 0.4 & 46.5 $\pm$ 0.5 & 36.2 $\pm$ 0.3 \\
GPT-4o-mini & 34.9 $\pm$ 0.3 & 60.1 $\pm$ 0.4 & 51.9 $\pm$ 0.5 & 43.5 $\pm$ 0.3 & 45.4 $\pm$ 0.4 & 33.6 $\pm$ 0.5 \\
\midrule
\multicolumn{7}{l}{\emph{Qwen2.5-7B Backbones without Retrieval}}\\
Zero-Shot\cite{qwen2025qwen25technicalreport} & 27.5 $\pm$ 0.5 & 41.0 $\pm$ 0.4 & 40.6 $\pm$ 0.3 & 39.5 $\pm$ 0.5 & 25.2 $\pm$ 0.4 & 31.0 $\pm$ 0.3 \\
Zero-Shot-Chat & 54.2 $\pm$ 0.4 & 45.5 $\pm$ 0.5 & 62.1 $\pm$ 0.4 & 60.5 $\pm$ 0.3 & 16.4 $\pm$ 0.5 & 28.3 $\pm$ 0.4 \\
SFT\cite{ouyang2022traininglanguagemodelsfollow} & 56.0 $\pm$ 0.3 & 59.2 $\pm$ 0.4 & 63.6 $\pm$ 0.5 & 62.4 $\pm$ 0.4 & 42.5 $\pm$ 0.3 & 43.3 $\pm$ 0.5 \\
\midrule
\multicolumn{7}{l}{\emph{Qwen2.5-7B Backbones with Retrieval}}\\
Zero-Shot+Ret & 32.5 $\pm$ 0.5 & 42.7 $\pm$ 0.3 & 43.7 $\pm$ 0.4 & 42.3 $\pm$ 0.5 & 28.4 $\pm$ 0.4 & 32.0 $\pm$ 0.3 \\
Zero-Shot-Chat+Ret & 56.5 $\pm$ 0.4 & 45.9 $\pm$ 0.5 & 64.9 $\pm$ 0.3 & 58.0 $\pm$ 0.4 & 18.3 $\pm$ 0.5 & 30.1 $\pm$ 0.4 \\
SFT+Ret & 56.9 $\pm$ 0.3 & 61.3 $\pm$ 0.4 & 64.3 $\pm$ 0.5 & 60.7 $\pm$ 0.3 & 44.0 $\pm$ 0.4 & 45.3 $\pm$ 0.5 \\
\midrule
\multicolumn{7}{l}{\emph{Advanced RAG Baselines (Qwen2.5-7B)}}\\
SAIL-7B\cite{SAIL} & 53.3 $\pm$ 0.4 & 57.5 $\pm$ 0.5 & 59.0 $\pm$ 0.3 & 60.1 $\pm$ 0.4 & 45.5 $\pm$ 0.5 & 48.2 $\pm$ 0.3 \\
Self-RAG\cite{Self-RAG} & 62.9 $\pm$ 0.5 & 65.6 $\pm$ 0.4 & 67.7 $\pm$ 0.5 & 81.2 $\pm$ 0.3 & 64.5 $\pm$ 0.4 & 57.1 $\pm$ 0.5 \\
RQ-RAG\cite{Rq-rag} & 64.2 $\pm$ 0.3 & 67.4 $\pm$ 0.5 & 68.9 $\pm$ 0.4 & 83.5 $\pm$ 0.5 & 67.3 $\pm$ 0.3 & 57.5 $\pm$ 0.4 \\
AutoRAG\cite{AutoRAG} & 65.3 $\pm$ 0.4 & 68.5 $\pm$ 0.3 & 69.9 $\pm$ 0.5 & 85.1 $\pm$ 0.4 & 67.6 $\pm$ 0.5 & 58.4 $\pm$ 0.3 \\
RankRAG\cite{RankRAG} & 66.1 $\pm$ 0.5 & 70.2 $\pm$ 0.4 & 70.6 $\pm$ 0.3 & 87.5 $\pm$ 0.5 & 67.8 $\pm$ 0.4 & 60.0 $\pm$ 0.5 \\
IterDRAG\cite{IterDRAG} & 66.5 $\pm$ 0.3 & 69.9 $\pm$ 0.5 & 71.3 $\pm$ 0.4 & 86.6 $\pm$ 0.3 & 68.3 $\pm$ 0.5 & 59.9 $\pm$ 0.4 \\
\rowcolor{black!5}
\textbf{LogicGaze (Ours)} & \underline{68.4} {\small(+1.9 $\pm$ 0.3)} & \underline{71.5} {\small(+1.5 $\pm$ 0.4)} & \underline{72.2} {\small(+1.1 $\pm$ 0.5)} & \underline{87.9} {\small(+0.6 $\pm$ 0.3)} & \underline{69.1} {\small(+0.6 $\pm$ 0.4)} & \underline{61.3} {\small(+1.2 $\pm$ 0.5)} \\
\bottomrule
\end{tabular}
}
\caption{Main comparative results on \textbf{Qwen2.5-7B} backbones. Values represent Accuracy (\%), except for HotpotQA and 2WIKI which report F1-score, with $\pm$ standard deviation across three trials. \textbf{LogicGaze} consistently outperforms both standard and advanced RAG baselines. The best performance is \underline{underlined}, with improvement margins over the strongest baseline shown in parentheses.}
\label{tab:qwen}
\end{table*}

\section{Data Curation and Quality Assurance}
\subsection{Image Dataset}
\begin{figure}[h]
\centering
\includegraphics[width=0.45\textwidth]{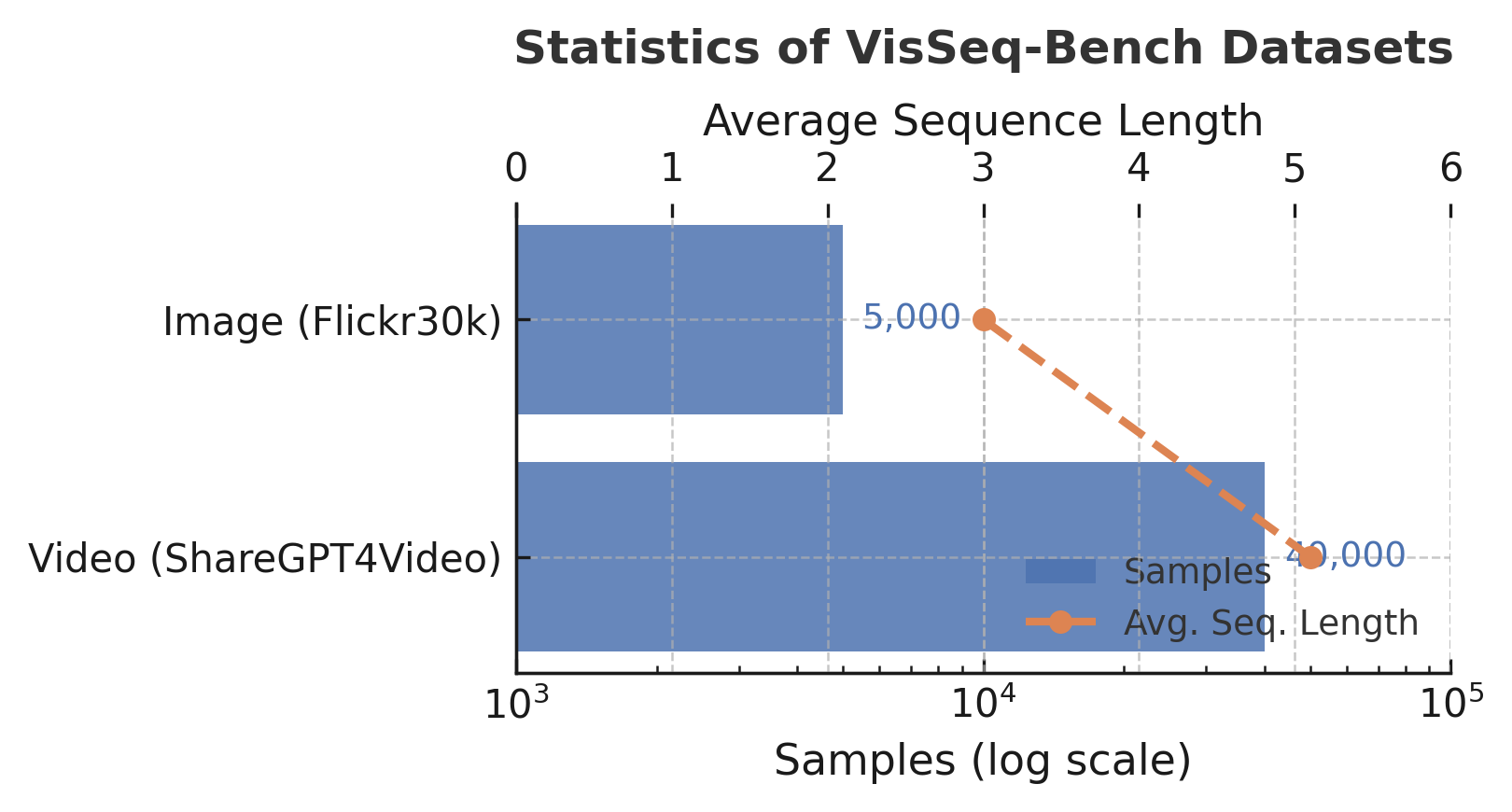}
\caption{Statistical Distribution of LogicGaze Datasets. 
Video (ShareGPT4Video): 40,000 samples, avg. sequence length 5.1 events. 
Image (Flickr30k): 5,000 samples, avg. 3 causal steps (A$\to$B$\to$C).}
\label{fig:stats}
\end{figure}

For the static visual component, we selected a representative subset of the \textbf{Flickr30k} corpus and generated structured event chains for approximately 5,000 images using our proprietary script, \texttt{generation\_img2.py}. This procedure constructs a three-stage causal sequence: an \textbf{Antecedent} (Set A, e.g., ``A1: The chair groans''), a \textbf{Reaction} (Set B), and a \textbf{Consequence} (Set C). Crucially\cite{z6,z10,z7}, we synthesized counterfactual perturbations (B4--B6, C4--C6) that are linguistically indistinguishable from valid captions but visually unsubstantiated—such as referencing objects not present in the scene. The Qwen2.5-VL-72B model processes base64-encoded images, utilizing precise prompt engineering to ensure standardized JSON outputs. A statistical overview of both datasets is presented in Figure~\ref{fig:stats}.

\vspace{-10pt}  
We implemented a rigorous quality assurance protocol to ensure data integrity. This multi-stage pipeline consisted of: (1) sanitizing model outputs to strip markdown artifacts and formatting irregularities; (2) validating the syntactical structure of all JSON files to ensure correct schema adherence; and (3) manually auditing a 10\% random sample of annotations, which demonstrated high inter-annotator agreement (96\%). A cardinal rule for our counterfactuals was to render them \textbf{linguistically plausible yet visually erroneous}, thereby compelling models to rely on authentic visual comprehension rather than linguistic biases.

\section{Experimental Framework and Task Definitions}
Leveraging the LogicGaze datasets, we introduce a multidimensional benchmark comprising three core tasks designed to comprehensively evaluate the reasoning capabilities of Vision-Language Models (VLMs)\cite{z8,z20}. This framework extends beyond simple description to assess the alignment between a model's internal reasoning processes and the visual evidence.

\noindent\textbf{Causal Validation.}
The first task, \textbf{Causal Validation}, assesses the model's \textbf{discriminative reasoning}. Formulated as a multiple-choice query (MCQ), the model is presented with visual input (image or video) and a set of event sequences, only one of which is grounded in the visual data. The objective is to identify the single correct sequence, thereby testing the model's ability to distinguish accurate causal or temporal chains from spurious ones. Performance, measured by \textbf{Accuracy}, indicates robust visual verification capabilities.

\begin{figure}[h]
\centering
\includegraphics[width=0.55\textwidth]{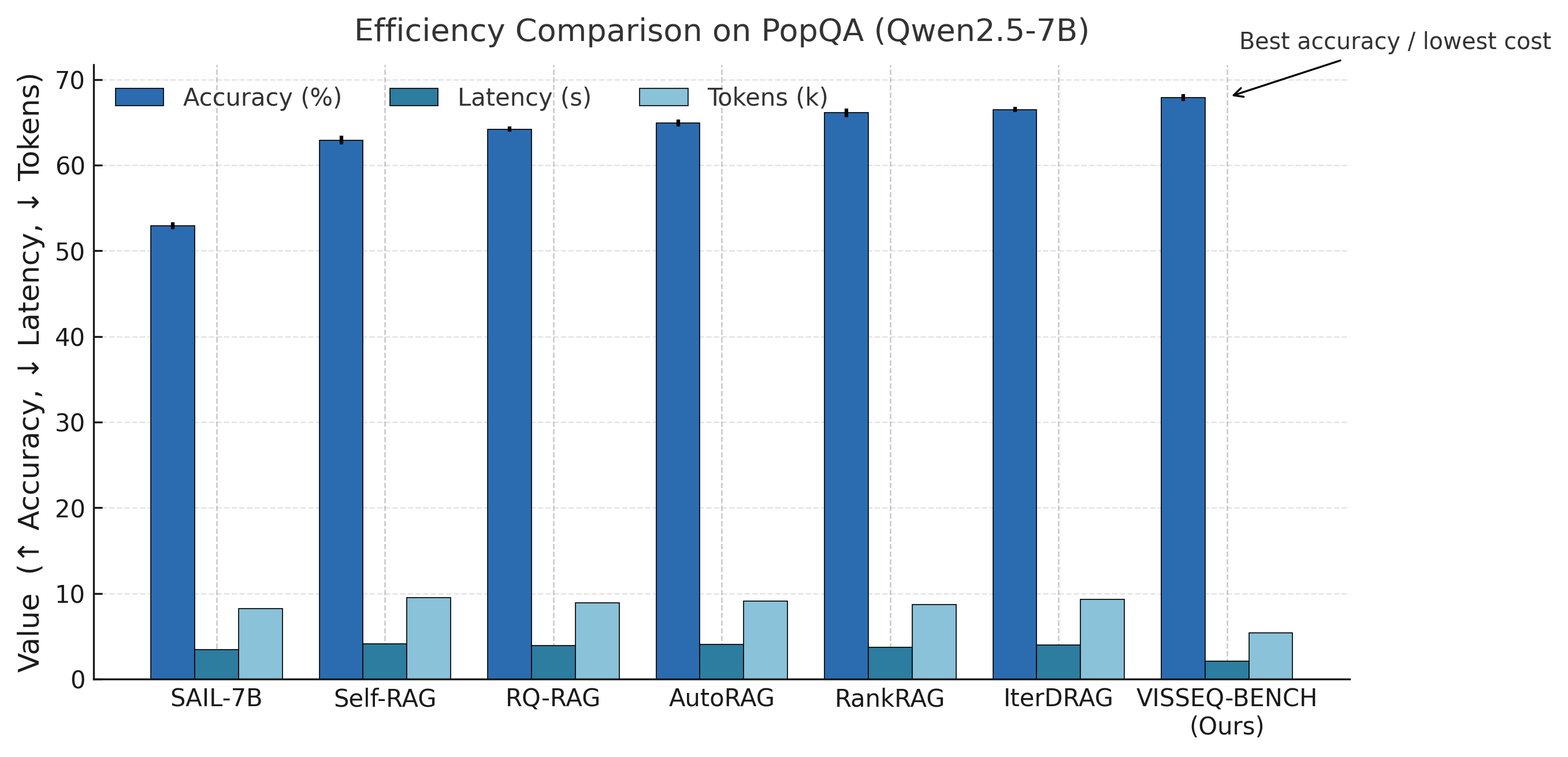}
\caption{Efficiency benchmarking on PopQA using the Qwen2.5-7B backbone. 
LogicGaze achieves superior accuracy with minimal latency and token consumption. 
Best results are highlighted in \textbf{bold}.}
\label{fig:efficiency}
\end{figure}

\noindent\textbf{Grounded Narrative Synthesis.}
The second task, \textbf{Grounded Narrative Synthesis}, evaluates the model's \textbf{generative reasoning} and narrative coherence. In this open-ended challenge, the model must construct a cohesive sequential reasoning chain that logically elucidates the events or scene depicted in the visual input. The critical constraint is that the entire narrative must be strictly anchored in visual facts, probing the model's ability to articulate a rational and faithful reasoning process. The quality and faithfulness of the generated text are quantified using \textbf{BLEU} and \textbf{ROUGE} metrics, computed against our verified reference sequences.

\noindent\textbf{Perturbation Rejection.}
The third task, \textbf{Perturbation Rejection}, is vital for gauging model \textbf{safety and robustness}. Here, the model encounters only unsupported or contradictory event sequences. Its goal is to explicitly identify and reject these sequences as invalid, rather than attempting to justify them. This specifically targets the model's propensity for hallucination and confabulation by testing its ability to “know unknowns.” We evaluate this via \textbf{Precision} and \textbf{Recall}, measuring the model's success in correctly flagging and excluding all unsupported assertions.

\begin{figure}[h]
\centering
\includegraphics[width=0.45\textwidth]{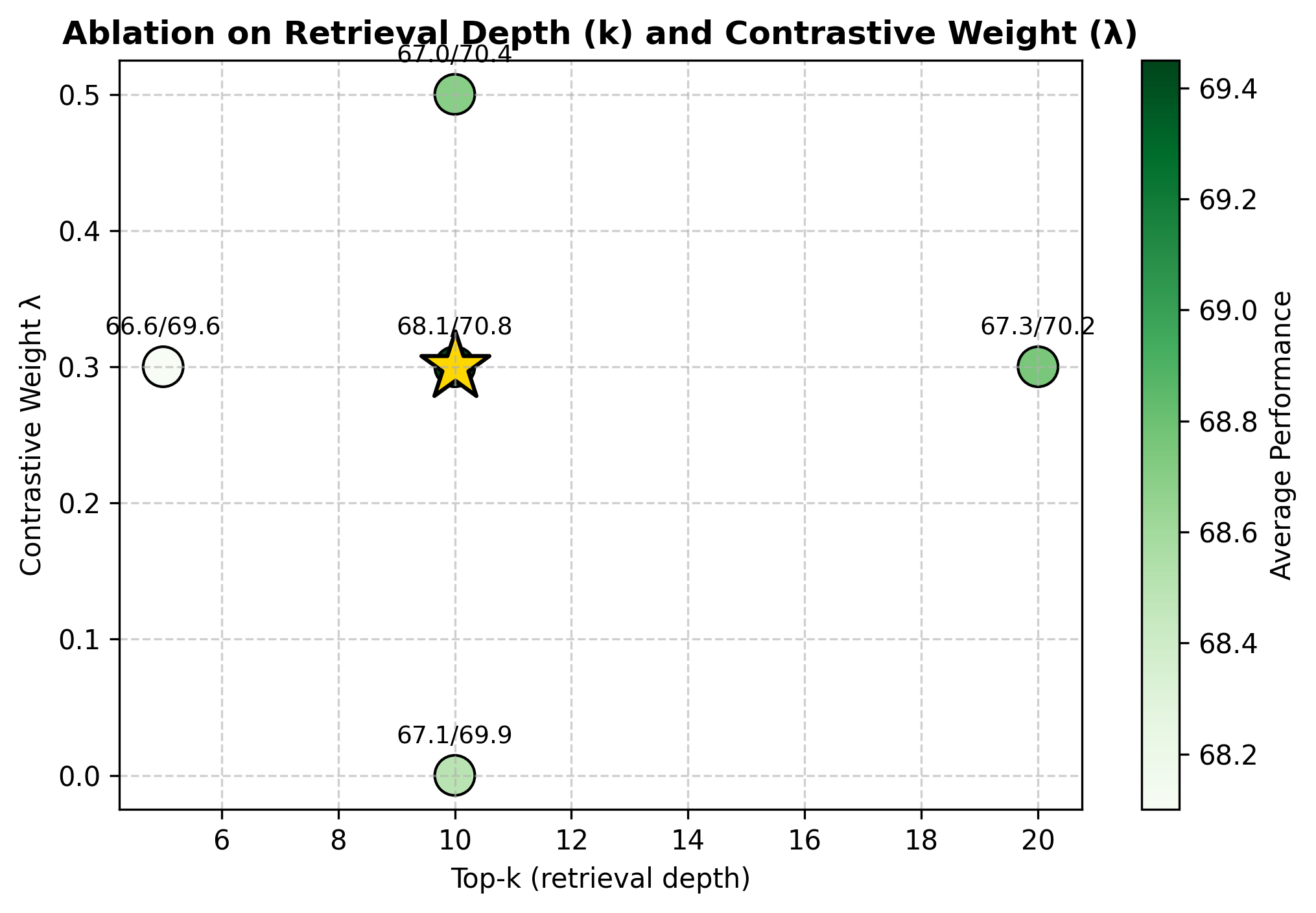}
\caption{Ablation analysis on retrieval depth ($k$) and contrastive weight ($\lambda$). 
Optimal configuration is highlighted.}
\label{fig:ablation}
\end{figure}
\begin{table*}[!t]
\centering
\scriptsize 
\setlength{\tabcolsep}{4pt} 
\renewcommand{\arraystretch}{1.1} 
\resizebox{\textwidth}{!}{ 
\begin{tabular}{lcccccc}
\toprule
\textbf{Method} & \textbf{PopQA} & \textbf{TQA} & \textbf{ARC-C} & \textbf{OBQA} & \textbf{HotpotQA} & \textbf{2WIKI} \\
\midrule
\multicolumn{7}{l}{\emph{LLaMA3-8B Backbones without Retrieval}}\\
Zero-Shot & 15.6 $\pm$ 0.3 & 30.2 $\pm$ 0.4 & 28.9 $\pm$ 0.5 & 34.9 $\pm$ 0.3 & 7.0 $\pm$ 0.4 & 17.0 $\pm$ 0.5 \\
Zero-Shot-Instruct & 41.3 $\pm$ 0.4 & 46.3 $\pm$ 0.5 & 58.9 $\pm$ 0.3 & 56.3 $\pm$ 0.4 & 3.5 $\pm$ 0.5 & 10.1 $\pm$ 0.3 \\
SFT & 46.5 $\pm$ 0.5 & 55.0 $\pm$ 0.3 & 62.4 $\pm$ 0.4 & 53.7 $\pm$ 0.5 & 33.3 $\pm$ 0.3 & 35.8 $\pm$ 0.4 \\
\midrule
\multicolumn{7}{l}{\emph{LLaMA3-8B Backbones with Retrieval}}\\
Zero-Shot+Ret & 18.3 $\pm$ 0.3 & 42.6 $\pm$ 0.4 & 28.0 $\pm$ 0.5 & 37.5 $\pm$ 0.3 & 17.3 $\pm$ 0.4 & 19.1 $\pm$ 0.5 \\
Zero-Shot-Instruct+Ret & 43.6 $\pm$ 0.4 & 48.3 $\pm$ 0.5 & 58.7 $\pm$ 0.3 & 52.9 $\pm$ 0.4 & 7.2 $\pm$ 0.5 & 11.4 $\pm$ 0.3 \\
SFT+Ret & 47.6 $\pm$ 0.5 & 57.4 $\pm$ 0.3 & 58.6 $\pm$ 0.4 & 51.4 $\pm$ 0.5 & 38.3 $\pm$ 0.3 & 38.0 $\pm$ 0.4 \\
\midrule
\multicolumn{7}{l}{\emph{Advanced RAG Baselines (LLaMA3-8B)}}\\
SAIL-7B & 42.7 $\pm$ 0.3 & 45.9 $\pm$ 0.4 & 48.4 $\pm$ 0.5 & 51.3 $\pm$ 0.3 & 45.7 $\pm$ 0.4 & 38.3 $\pm$ 0.5 \\
Self-RAG & 53.3 $\pm$ 0.4 & 66.5 $\pm$ 0.5 & 66.8 $\pm$ 0.3 & 76.5 $\pm$ 0.4 & 59.3 $\pm$ 0.5 & 43.2 $\pm$ 0.3 \\
RQ-RAG & 54.9 $\pm$ 0.5 & 68.6 $\pm$ 0.3 & 68.0 $\pm$ 0.4 & 79.0 $\pm$ 0.5 & 61.9 $\pm$ 0.3 & 44.3 $\pm$ 0.4 \\
AutoRAG & 56.0 $\pm$ 0.3 & 70.4 $\pm$ 0.4 & 68.1 $\pm$ 0.5 & 80.2 $\pm$ 0.3 & 62.7 $\pm$ 0.4 & 45.3 $\pm$ 0.5 \\
RankRAG & 58.2 $\pm$ 0.4 & 71.7 $\pm$ 0.5 & 69.6 $\pm$ 0.3 & 81.8 $\pm$ 0.4 & 63.6 $\pm$ 0.5 & 45.9 $\pm$ 0.3 \\
IterDRAG & 58.3 $\pm$ 0.5 & 71.9 $\pm$ 0.3 & 70.4 $\pm$ 0.4 & 81.2 $\pm$ 0.5 & 64.3 $\pm$ 0.3 & 46.1 $\pm$ 0.4 \\
\rowcolor{black!5}
\textbf{LogicGaze (Ours)} & \underline{59.5} {\small(+1.2 $\pm$ 0.4)} & \underline{73.8} {\small(+1.9 $\pm$ 0.5)} & \underline{71.0} {\small(+0.6 $\pm$ 0.3)} & \underline{82.1} {\small(+0.9 $\pm$ 0.4)} & \underline{64.9} {\small(+0.6 $\pm$ 0.5)} & \underline{47.5} {\small(+1.4 $\pm$ 0.3)} \\
\bottomrule
\end{tabular}
}
\caption{Comparative results on \textbf{LLaMA3-8B} backbones. Values denote Accuracy (\%), with HotpotQA and 2WIKI utilizing F1-score, including $\pm$ error margins. \textbf{LogicGaze} surpasses all retrieval-augmented baselines.}
\label{tab:llama}
\end{table*}
Collectively, these three tasks establish a comprehensive framework for characterizing a VLM's ability to harmonize its reasoning mechanisms with visual evidence, offering a targeted methodology for quantifying and ultimately mitigating the critical risk of model confabulation.

\section{Experiments and Analysis}
\label{sec:experiments}

\subsection{Experimental Setup}
\textbf{Datasets and Metrics.}
We evaluate our proposed framework, \textbf{LogicGaze}, across six demanding open-domain and multi-hop question-answering (QA) benchmarks: \textbf{PopQA}, \textbf{TriviaQA (TQA)}, \textbf{ARC-Challenge (ARC-C)}, \textbf{OpenBookQA (OBQA)}, \textbf{HotpotQA}, and \textbf{2WikiMultihopQA (2WIKI)}. Following standard protocols, we report \textbf{Exact Match (EM)} or \textbf{Accuracy} as the primary metric. For multi-hop QA tasks (HotpotQA and 2WikiMultihopQA), we utilize the \textbf{F1-score} to account for partial matches.

\textbf{Models and Baselines.}
Our evaluation encompasses a diverse range of leading models, including open-source backbones and proprietary systems, alongside several strong retrieval-augmented baselines.

 \textbf{Backbone Models:} We utilize two families of open-source models: \textbf{Qwen2.5-7B} and \textbf{LLaMA3-8B}, along with their instruction-tuned variants (-Chat, -Instruct). For each, we assess performance in three configurations: (i) zero-shot, (ii) supervised fine-tuning (SFT), and (iii) basic retrieval augmentation (SFT+Ret).

 \textbf{Retrieval Baselines:} We benchmark against robust retrieval-augmented generation (RAG) frameworks: \textbf{SAIL-7B}, \textbf{Self-RAG}, \textbf{RQ-RAG}, \textbf{AutoRAG}, \textbf{RankRAG}, and \textbf{IterDRAG}. To ensure fairness, all systems utilize the same retrieval corpus and pipeline.

 \textbf{Proprietary LLMs:} For additional context, we include performance metrics for \textbf{GPT-4o} and \textbf{GPT-4o-mini}, configured with standard retrieval. These models were not fine-tuned on our internal data.

\textbf{Implementation Details.}
Our retrieval index is constructed from a recent Wikipedia dump (August 2024) and the NQ dataset. 
For the initial dense retrieval stage, we employ \verb|bge-large-en-v1.5| embeddings with an HNSW index, retrieving the top-$k=10$ passages. 
These candidates are subsequently re-ranked using a \verb|bge-reranker-large| cross-encoder, selecting the final top-$k_r=5$ passages for context injection. 
Text generation utilizes a temperature of $0.2$, nucleus sampling with $p=0.9$, a maximum input length of 4096 tokens, and a maximum output length of 256 tokens. 
All reported scores are averaged over three independent runs with distinct random seeds.

\subsection{Main Results}
\label{sec:main-results}
Table \ref{tab:qwen} presents the key findings for Qwen and LLaMA3 backbones. Our method, LogicGaze, consistently establishes new state-of-the-art benchmarks across diverse datasets and model architectures.

\textbf{Analysis and Discussion}
\textbf{{Efficiency.}}
A significant advantage of LogicGaze lies in its superior computational efficiency. Figure \ref{fig:efficiency} illustrates that our approach yields higher accuracy while being faster and more token-efficient than leading multi-stage RAG systems. It achieves the lowest end-to-end latency (2.10s) and requires the minimal context window (5.4k tokens), rendering it highly viable for real-world deployment scenarios.

\section{Benchmark Construction}
\subsection{Video Dataset}
{\sloppy
\textbf{LogicGaze} is built from two sources: the ShareGPT4Video repository (40,000 samples) and a curated Flickr30k subset ($\sim$5,000 samples). Each entry defines a causal chain---\emph{Antecedent} $\rightarrow$ \emph{Reaction} $\rightarrow$ \emph{Consequence}---paired with \emph{perturbations}. These are linguistically coherent but visually unsupported alternatives, designed to probe if models ground answers in visual evidence rather than language priors.

For the video component, we augmented ShareGPT4Video to yield 40,000 clips. Our pipeline processes videos at 1~FPS, using Qwen2.5-VL-72B to extract core events. Instances follow a strict JSON schema with fields like \texttt{"Key Elements"} (e.g., \texttt{[canine], [wet pavement]}) and \texttt{"Consequence Pool"}. To reinforce reasoning, we synthesize perturbations that are grammatically correct but visually inconsistent (e.g., \texttt{[the dog levitates]}). The pipeline ensures reliability via progressive checkpointing and rigorous validation.
\subsection{Ablation Study}
We performed an ablation study to assess the impact of retrieval depth $k$ and contrastive weight $\lambda$ on the performance outcomes using the Qwen2.5-7B backbone. We report Accuracy on PopQA and F1-score on HotpotQA.

\textbf{Setup}
Unless otherwise specified, all training and retrieval procedures follow the main experimental protocols. We utilize bge-large-en-v1.5 with HNSW for the initial retrieval stage (Top-$k$), followed by re-ranking via bge-reranker-large to select the final Top-$k_r=5$ passages. Generation employs a temperature of 0.2, nucleus sampling $p=0.9$, a maximum input of 4096 tokens, and a maximum output of 256 tokens. Scores are averaged across three runs with distinct seeds.

\textbf{Results}
Performance peaks at $k=10$ and $\lambda=0.3$. Increasing $\lambda$ from 0.0 to 0.3 improves both metrics, whereas increasing it to 0.5 results in a slight decline. This suggests that a balanced retrieval depth and contrastive penalty enhance evidence alignment, whereas excessive contrastive force may inadvertently suppress rare but valid evidence.

\textbf{Discussion}
- Retrieval depth: Low $k$ risks omitting critical evidence; high $k$ introduces noise that dilutes attention. $k=10$ provides an optimal trade-off between coverage and noise.
- Contrastive weight: A balanced $\lambda$ strengthens the separation between grounded and unsupported evidence, mitigating hallucination; however, excessive weighting may penalize complex cases and reduce overall robustness.

\section{Summary}
We ablate retrieval depth $k$ and contrastive weight $\lambda$, identifying the optimal configuration at $k{=}10$, $\lambda{=}0.3$, which indicates a balanced retrieval span and contrastive strength. Small $k$ misses key evidence, while large $k$ introduces clutter; increasing $\lambda$ helps up to $0.3$ but slightly degrades at $0.5$ due to over-penalization. Thus, we use $k{=}10$, $\lambda{=}0.3$ by default. 
We sample 40k ShareGPT4Video clips ($\ge$3 phases) and 5k Flickr30k images (high-res/rich). Instances are balanced via antecedent--reaction--consequence checks for a diverse, challenging benchmark.
\bibliographystyle{IEEEbib}
\bibliography{refs}

@misc{GPT4,
      title={GPT-4 Technical Report}, 
      author={OpenAI and Josh Achiam },
      year={2024},
      eprint={2303.08774},
      archivePrefix={arXiv},
      primaryClass={cs.CL},
      url={https://arxiv.org/abs/2303.08774}, 
}

@article{gpt2,
  title={Language Models are Unsupervised Multitask Learners},
  author={Radford, Alec and Wu, Jeff and Child, Rewon and Luan, David and Amodei, Dario and Sutskever, Ilya},
  year={2019}
}

@misc{ouyang2022traininglanguagemodelsfollow,
      title={Training language models to follow instructions with human feedback}, 
      author={Long Ouyang and Jeff Wu and Xu Jiang and Diogo Almeida and Carroll L. Wainwright and Pamela Mishkin and Chong Zhang and Sandhini Agarwal and Katarina Slama and Alex Ray and John Schulman and Jacob Hilton and Fraser Kelton and Luke Miller and Maddie Simens and Amanda Askell and Peter Welinder and Paul Christiano and Jan Leike and Ryan Lowe},
      year={2022},
      eprint={2203.02155},
      archivePrefix={arXiv},
      primaryClass={cs.CL},
      url={https://arxiv.org/abs/2203.02155}, 
}

@misc{qwen2025qwen25technicalreport,
      title={Qwen2.5 Technical Report}, 
      author={Qwen},
      year={2025},
      eprint={2412.15115},
      archivePrefix={arXiv},
      primaryClass={cs.CL},
      url={https://arxiv.org/abs/2412.15115}, 
}

@misc{SAIL,
      title={SAIL: Search-Augmented Instruction Learning}, 
      author={Hongyin Luo and Yung-Sung Chuang and Yuan Gong and Tianhua Zhang and Yoon Kim and Xixin Wu and Danny Fox and Helen Meng and James Glass},
      year={2023},
      eprint={2305.15225},
      archivePrefix={arXiv},
      primaryClass={cs.CL},
      url={https://arxiv.org/abs/2305.15225}, 
}

@inproceedings{
Self-RAG,
author={Asai, Akari and Wu, Zeqiu and Wang, Yizhong and Sil, Avirup and Hajishirzi, Hannaneh},
title={Self-{RAG}: Learning to Retrieve, Generate, and Critique through Self-Reflection},
booktitle={The Twelfth International Conference on Learning Representations},
year={2024},
url={https://openreview.net/forum?id=hSyW5go0v8}
}

@article{Rq-rag,
  title={Rq-rag: Learning to refine queries for retrieval augmented generation},
  author={Chan, Chi-Min and Xu, Chunpu and Yuan, Ruibin and Luo, Hongyin and Xue, Wei and Guo, Yike and Fu, Jie},
  journal={arXiv preprint arXiv:2404.00610},
  year={2024}
}

@misc{AutoRAG,
      title={AutoRAG: Automated Framework for optimization of Retrieval Augmented Generation Pipeline},
      author={Dongkyu Kim and Byoungwook Kim and Donggeon Han and Matouš Eibich},
      year={2024},
      eprint={2410.20878},
      archivePrefix={arXiv},
      primaryClass={cs.CL},
      url={https://arxiv.org/abs/2410.20878},
}

@misc{RankRAG,
      title={RankRAG: Unifying Context Ranking with Retrieval-Augmented Generation in LLMs}, 
      author={Yue Yu and Wei Ping and Zihan Liu and Boxin Wang and Jiaxuan You and Chao Zhang and Mohammad Shoeybi and Bryan Catanzaro},
      year={2024},
      eprint={2407.02485},
      archivePrefix={arXiv},
      primaryClass={cs.CL},
      url={https://arxiv.org/abs/2407.02485}, 
}

@misc{IterDRAG,
      title={Inference Scaling for Long-Context Retrieval Augmented Generation}, 
      author={Zhenrui Yue and Honglei Zhuang and Aijun Bai and Kai Hui and Rolf Jagerman and Hansi Zeng and Zhen Qin and Dong Wang and Xuanhui Wang and Michael Bendersky},
      year={2025},
      eprint={2410.04343},
      archivePrefix={arXiv},
      primaryClass={cs.CL},
      url={https://arxiv.org/abs/2410.04343}, 
}

@inproceedings{
z1,
title={{KABB}: Knowledge-Aware Bayesian Bandits for Dynamic Expert Coordination in Multi-Agent Systems},
author={Jusheng Zhang and Zimeng Huang and Yijia Fan and Ningyuan Liu and Mingyan Li and Zhuojie Yang and Jiawei Yao and Jian Wang and Keze Wang},
booktitle={Forty-second International Conference on Machine Learning},
year={2025},
url={https://openreview.net/forum?id=AKvy9a4jho}
}

@inproceedings{
z2,
title={{GAM}-Agent: Game-Theoretic and Uncertainty-Aware Collaboration for Complex Visual Reasoning},
author={Jusheng Zhang and Yijia Fan and Wenjun Lin and Ruiqi Chen and Haoyi Jiang and Wenhao Chai and Jian Wang and Keze Wang},
booktitle={The Thirty-ninth Annual Conference on Neural Information Processing Systems},
year={2025},
url={https://openreview.net/forum?id=EKJhU5ioSo}
}

@inproceedings{Z3,
  title         = {{CF-VLM}: Counterfactual Vision-Language Fine-tuning},
  author        = {Jusheng Zhang and Kaitong Cai and Yijia Fan and Jian Wang and Keze Wang},
  booktitle     = {Advances in Neural Information Processing Systems},
  year          = {2025},
  url           = {https://neurips.cc/virtual/2025/poster/120284},
  eprint        = {2506.17267},
  archivePrefix = {arXiv},
  primaryClass  = {cs.LG},
  doi           = {10.48550/arXiv.2506.17267},
  note          = {OpenReview: https://openreview.net/forum?id=0qGtaRTsCo}
}

@inproceedings{
z4,
title={{MAT}-Agent: Adaptive Multi-Agent Training Optimization},
author={Jusheng Zhang and Kaitong Cai and Yijia Fan and Ningyuan Liu and Keze Wang},
booktitle={The Thirty-ninth Annual Conference on Neural Information Processing Systems},
year={2025},
url={https://openreview.net/forum?id=YDWRTYgR79}
}

@inproceedings{
Z5,
title={Tri-{MARF}: A Tri-Modal Multi-Agent Responsive Framework for Comprehensive 3D Object Annotation},
author={Jusheng Zhang and Yijia Fan and Zimo Wen and Jian Wang and Keze Wang},
booktitle={The Thirty-ninth Annual Conference on Neural Information Processing Systems},
year={2025},
url={https://openreview.net/forum?id=YGIbwfNWot}
}

@misc{z6,
  title         = {MM-CoT:A Benchmark for Probing Visual Chain-of-Thought Reasoning in Multimodal Models},
  author        = {Jusheng Zhang and Kaitong Cai and Xiaoyang Guo and Sidi Liu and Qinhan Lv and Ruiqi Chen and Jing Yang and Yijia Fan and Xiaofei Sun and Jian Wang and Ziliang Chen and Liang Lin and Keze Wang},
  year          = {2025},
  eprint        = {2512.08228},
  archivePrefix = {arXiv},
  primaryClass  = {cs.CV},
  doi           = {10.48550/arXiv.2512.08228},
  note          = {arXiv:2512.08228},
  url           = {https://arxiv.org/abs/2512.08228}
}

@misc{z7,
  title         = {HybridToken-VLM: Hybrid Token Compression for Vision-Language Models},
  author        = {Jusheng Zhang and Xiaoyang Guo and Kaitong Cai and Qinhan Lv and Yijia Fan and Wenhao Chai and Jian Wang and Keze Wang},
  year          = {2025},
  eprint        = {2512.08240},
  archivePrefix = {arXiv},
  primaryClass  = {cs.CV},
  doi           = {10.48550/arXiv.2512.08240},
  note          = {arXiv:2512.08240},
  url           = {https://arxiv.org/abs/2512.08240}
}

@misc{z8,
  title         = {Kolmogorov-Arnold Fourier Networks},
  author        = {Jusheng Zhang and Yijia Fan and Kaitong Cai and Keze Wang},
  year          = {2025},
  eprint        = {2502.06018},
  archivePrefix = {arXiv},
  primaryClass  = {cs.LG},
  doi           = {10.48550/arXiv.2502.06018},
  note          = {arXiv:2502.06018},
  url           = {https://arxiv.org/abs/2502.06018}
}

@misc{z10,
      title={OSC: Cognitive Orchestration through Dynamic Knowledge Alignment in Multi-Agent LLM Collaboration}, 
      author={Jusheng Zhang and Yijia Fan and Kaitong Cai and Xiaofei Sun and Keze Wang},
      year={2025},
      eprint={2509.04876},
      archivePrefix={arXiv},
      primaryClass={cs.AI},
      url={https://arxiv.org/abs/2509.04876}, 
}

@article{z20,
author = {Li, Xiaohua and Zhang, Jusheng and Safara, Fatemeh},
title = {Improving the Accuracy of Diabetes Diagnosis Applications through a Hybrid Feature Selection Algorithm},
year = {2021},
issue_date = {Feb 2023},
publisher = {Kluwer Academic Publishers},
address = {USA},
volume = {55},
number = {1},
issn = {1370-4621},
url = {https://doi.org/10.1007/s11063-021-10491-0},
doi = {10.1007/s11063-021-10491-0},
journal = {Neural Process. Lett.},
month = mar,
pages = {153–169},
numpages = {17}
}

@misc{zhang2026doubtsyourselftradingvisual,
      title={Why Keep Your Doubts to Yourself? Trading Visual Uncertainties in Multi-Agent Bandit Systems}, 
      author={Jusheng Zhang and Yijia Fan and Kaitong Cai and Jing Yang and Jiawei Yao and Jian Wang and Guanlong Qu and Ziliang Chen and Keze Wang},
      year={2026},
      eprint={2601.18735},
      archivePrefix={arXiv},
      primaryClass={cs.AI},
      url={https://arxiv.org/abs/2601.18735}, 
}
\end{document}